# Leveraging dendritic properties to advance machine learning and neuro-inspired computing

*Michalis Pagkalos[1,2], Roman Makarov[1,2], and Panayiota Poirazi[1,*]*

[1]Institute of Molecular Biology and Biotechnology (IMBB), Foundation for Research and Technology Hellas (FORTH), Heraklion, 70013, Greece

[2]Department of Biology, University of Crete, Heraklion, 70013, Greece

[*]Corresponding author: poirazi@imbb.forth.gr

## Highlights

- There is urging demand for efficient and sustainable artificial intelligence systems
- Brain-inspired computing is a promising way forward
- Dendrites are an indispensable component of biological intelligence
- Dendro-inspired algorithms offer compelling and unique advantages
- Dendritic mechanisms have established applications in neuromorphic systems

## Abstract

The brain is a remarkably capable and efficient system. It can process and store huge amounts of noisy and unstructured information using minimal energy. In contrast, current artificial intelligence (AI) systems require vast resources for training while still struggling to compete in tasks that are trivial for biological agents. Thus, brain-inspired engineering has emerged as a promising new avenue for designing sustainable, next-generation AI systems. Here, we describe how dendritic mechanisms of biological neurons have inspired innovative solutions for significant AI problems, including credit assignment in multilayer networks, catastrophic forgetting, and high energy consumption. These findings provide exciting alternatives to existing architectures, showing how dendritic research can pave the way for building more powerful and energy-efficient artificial learning systems.

# Introduction

Artificial intelligence (AI) has experienced remarkable growth in recent years, promising to revolutionize various aspects of our society and everyday lives. However, the long-term sustainability of AI progress has surfaced as a major concern [1–3]. Current state-of-the-art AI systems rely heavily on parameter scaling to improve performance, resulting in an insatiable demand for resources [4]. Meanwhile, the chip manufacturing industry is struggling to keep up with our ever-increasing need to handle the tremendous amount of data we generate daily [4–6]. Therefore, it is imperative to explore alternative approaches to improve both our algorithms and hardware, ensuring sustainable progress in the long run.

Neuro-inspired engineering has emerged as a promising avenue for developing the computing systems of tomorrow by mimicking the operation and efficiency of the human brain [2,3,7,8]. Among the various components of neural architecture, dendrites have recently gained considerable attention as a source of inspiration for designing novel algorithms and hardware architectures [9–11]. Similar to neuronal axons, dendrites possess voltage-gated ion channels that allow them to generate highly heterogeneous regenerative events called dendritic spikes (dSpikes). Thanks to dSpikes, dendrites can operate as highly non-linear, semi-independent integration units, greatly expanding the range of computations a single neuron can perform [12–14]. Additionally, dendritic mechanisms have been causally linked to crucial brain functions, including sensory perception, plasticity, and behavior [15,16].

This review summarizes tangible examples of how dendritic mechanisms have inspired the development of innovative network architectures and training algorithms to address important AI problems. Specifically, we focus on three compelling applications: a) Learning in multilayer neural networks, b) Mitigating the effects of catastrophic forgetting, and c) Developing efficient algorithms compatible with neuro-inspired hardware. By exploring these applications, we aim to demonstrate the great potential of dendritic research to revolutionize the field of machine learning and neuromorphic engineering.

# Learning in multilayer networks

Deep neural networks (DNNs) are artificial neural networks that consist of multiple layers of interconnected processing units (neurons) organized hierarchically (**Figure 1a**). When DNNs are trained on a new task, they need to determine how to adjust specific network parameters (e.g., the synaptic weights) to respond to some input data with a desired output. A standard solution to this problem, also known as the "credit assignment problem," is the error-backpropagation algorithm (backprop), which performs two primary functions. First, it calculates the error (difference) between a network prediction and a target output at every simulation step. Then, it attempts to minimize that error by updating the trainable parameters in all network layers, starting from the last layer and moving backward.

While backprop is currently the most widely-used training algorithm in deep learning, it has several drawbacks [17]. Firstly, it is inefficient as it calculates error gradients for the entire network at every simulation iteration. Secondly, it relies on a symmetric feedback structure to communicate error gradients to previous layers, which is biologically implausible. Thirdly, it requires substantial amounts of data and multiple training sessions to achieve optimal performance. Finally, it is not well-suited for unsupervised learning when the target output is unknown (unlabeled data). Given that the brain consists of multilayer networks, capable of supervised and unsupervised learning while utilizing only locally available information and a few training examples, an intriguing question arises: How has nature solved the credit assignment problem?

Recent modeling studies, inspired by experimental [18,19] and early theoretical research [20], demonstrate that learning in DNNs is possible when certain biological constraints are considered. Although these studies diverge in their specific predictions, they use similar principles inspired by how dendrites integrate synaptic information (**Figure 1b**). Importantly, they extend the point-neuron model by adding dendritic compartments, allowing a single neuron to integrate multiple, specially segregated input pathways. In this way, feedforward (sensory / input) information can be combined with feedback (high-order / context) signals coming from different network layers. This layer-specific pathway interaction can generate non-linear somato-dendritic events that allow the calculation of local errors, which can then coordinate plasticity and learning across all layers.

Along these lines, Gergiev et al. (2017) [17] described a learning rule that relies on the dendritic responses generated by feedback signals to the apical dendrites of the previous layers (**Figure 1c**). In this example, learning is achieved by minimizing the difference between dendritic plateaus produced during a "target" and a "forward" phase, where a teacher input to the output layer is present or absent, respectively. By contrast, Sacramento et al. (2018) [21] proposed that dendritic plateaus can directly serve as error signals that need to be eliminated (linearized) through dendrite-targeting lateral inhibition (**Figure 1d**). One advantage of this approach is that the network can operate continuously, without requiring two separate temporal phases, as in Geurguiev et al. (2017) [17].

More recently, Payeur et al. (2021) [22] first presented how active dendritic mechanisms, combined with a somatic, burst-dependent learning rule, enable sophisticated credit assignment in DNNs (**Figure 1e**). In this model, feedback inputs to the apical dendrites can induce the generation of dendritic plateau potentials, which in turn cause neurons to fire with high-frequency bursts. The latter can serve as powerful instructive signals for plasticity, allowing learning across all network layers. This study built upon previous elegant work on multiplexing [23], showing that active dendrites enable neuronal ensembles to simultaneously communicate multiple information streams, encoded in different neuronal output statistics (**Figure 1f**). Notably, the multiplexing theory provides a theoretical basis of how dendritic mechanisms can increase the amount of information that can be transmitted by a fixed number of axons in the brain. Finally, recent work by Greedy et al. (2022) [24], combined the above ideas (dendritic multiplexing, a burst-dependent learning rule, and dendritic inhibition) into a single model, namely *Bursting Cortico-Cortical Networks* (BurstCCN). BurstCCN was shown to be highly effective in training multilayer architectures, while only requiring a single-phase learning process.

The above studies highlight the potential of dendrites as a valuable source of inspiration for developing effective training algorithms for supervised learning in DNNs. However, biological agents can also learn from unlabeled examples, or when a target behavior is unknown (unsupervised learning). In that regard, a novel Hebbian rule, namely *Contrastive Local And Predictive Plasticity* (CLAPP) has been proposed [25]. CLAPP, among other biological properties, accounts for the influence of dendritic activity on synaptic plasticity. Dendrite-targeting synapses (both lateral and feedback), can serve as predictive signals of neuronal activity that bias weight updates in the entire network. Interestingly, this study aligns with the latest experimental and theoretical research, suggesting that dendrites may have a prominent role in coordinating efficient hierarchical predictive coding in the cortex [19,26].

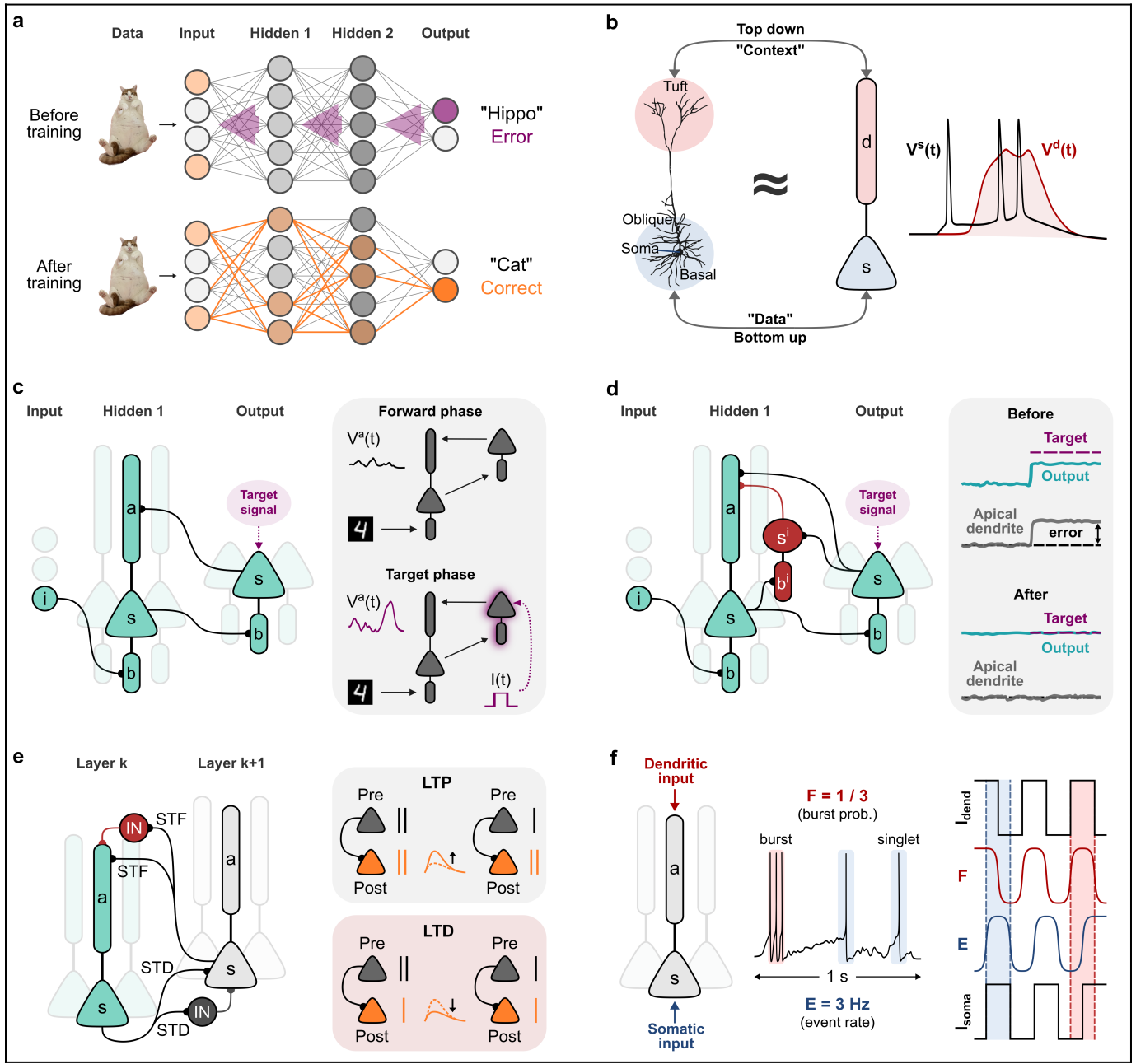

**Figure 1 | Training deep neural networks with dendro-inspired learning rules.**

**a)** Illustration of a DNN trained for image classification using standard backpropagation of errors. <u>Top</u>: Before training, the network fails to correctly classify a cat's image. This generates error signals (purple arrows) at the output layer that are sequentially transmitted to the previous layers. <u>Bottom</u>: The error signals are used to update the synaptic weights (thick orange lines) that activate a population of neurons (orange nodes) that will give the correct answer.

**b)** Biological neurons are far more complex than the point-neuron models commonly used in deep learning. Cortical pyramidal neurons have at least two functional dendritic domains. A perisomatic integration zone (blue circle) receives feedforward inputs, and a distal integration zone (red circle) receives contextual signals from other areas. The voltage traces on the right illustrate how dendritic events can modulate the frequency of the somatic output.

**c)** Description of the model in Guerguiev et al. (2017). <u>Left</u>: Schematic of the network. The hidden layers consist of three-compartment neurons (soma, apical, basal) and the output layer of two-compartmental neurons (soma, basal). Feedforward synapses project to basal dendrites of the next layer. Feedback synapses project to the apical dendrites of the previous layer. A target signal (current pulse) can be directly provided to the soma of an output neuron. <u>Right</u>: Illustration of a two-phase learning rule. During the forward phase, no target signal is provided to the network. During the target phase, a target signal stimulates the "correct" output neuron, triggering the generation of dendritic plateaus at the previous layer(s).

This rule aims to minimize the difference between plateaus generated in the presence or absence of teacher signals by updating the feedforward synaptic weights.

**d)** Description of the model in Sacramento et al. (2018). <u>Left</u>: Schematic of the network. The morphological characteristics of the excitatory neurons, along with the distribution of excitatory synapses, are as in Guerguiev et al. (2017). However, this model also includes two-compartmental interneurons (red) that target the apical dendrites of same-layer neurons. The interneurons receive feedforward input at their basal compartment and feedback input from the next layer at their somatic compartment. <u>Right</u>: Indicative voltage responses before and after a new teaching signal is learned. Before learning, a teacher signal at the output layer triggers dendritic activity at the previous layer(s). The goal of the learning rule here is to linearize the dendritic responses through lateral inhibition by updating the feedforward synaptic weights.

**e)** Description of the model in Payeur et al. (2021). <u>Left</u>: Schematic of the network. The network consists of two-compartmental (soma, apical) excitatory neurons and point interneurons. Lower-level neurons (cyan) project with short-term depressing (STD) synapses to the somatic compartments of higher-level neurons (light gray) and to inhibitory neurons (dark gray), providing disynaptic, somatic inhibition. Higher-level neurons project with short-term facilitating (STF) synapses to the dendritic compartments of lower-level neurons and to inhibitory neurons (red), providing disynaptic dendritic inhibition. STD and STF synapses can communicate "event" and "burst" signals, respectively (also see **panel f**). <u>Right</u>: Schematic of the learning rule. When presynaptic activity leads to postsynaptic bursting, it results in synaptic potentiation (top), whereas isolated postsynaptic spikes lead to synaptic depression.

**f)** Illustration of how neurons with active dendrites can communicate multiple information streams encoded in different output statistics. Neuronal activity can be classified as isolated spike events (singlets) or burst events. In this example, the neuron fires a burst and two singlets within 1 second. <u>Right</u>: Dendritic input is represented by the bursting probability (F), whereas somatic input is represented by the event rate (E).

## A dendritic solution to catastrophic forgetting

In deep learning, "catastrophic forgetting" is a phenomenon where a neural network rapidly forgets a previously learned task (e.g., task A), when trained on a new task (e.g., task B). This happens because the network parameters optimized for task A (using standard backprop), can be altered to meet the specific demands of task B (**Figure 2a**). This is a major issue because it significantly limits the number of tasks a single network can store, hindering its computational capacity. Interestingly, recent experimental and theoretical studies highlight that dendrites could offer a natural remedy for catastrophic forgetting.

Multiple lines of evidence suggest that individual dendritic branches, rather than neurons, serve as the fundamental unit of plasticity and learning in the mammalian brain. Dendrites are crucial in mitigating catastrophic forgetting by enabling selective updates of important synapses for a new task while leaving other synapses unchanged. For example, Cichon and Gan (2015) [27] demonstrated *in vivo* that different motor learning tasks trigger dendritic spiking and long-term potentiation of synapses in non-overlapping dendrites, thereby reducing memory interference. Inspired by such findings, two independent studies, Kirkpatrick et al. (2017) [28] and Zenke et al. (2017) [29], proposed Elastic Weight Consolidation (EWC) and Synaptic Intelligence (SI), respectively, two algorithms that differ in their mathematical implementation, but share the same goal. Both algorithms identify the synaptic weights that are the most important for a learned task and make them less plastic when a new task is learned. As a result, EWC and SI enable DNNs of fixed size to sequentially learn multiple tasks without forgetting the previously learned ones. Recent modeling studies by Kastellakis et al. (2016) [30], Bono and Clopath (2017) [31], and Limbacher and Legenstein (2020) [32] have provided a mechanistic explanation of how biological neurons implement similar algorithms. According to these studies, a combination of NMDA-mediated plasticity and clustering of temporally and functionally related synapses can result in memory stabilization and sparse ensemble storage.

Another way dendrites may help to mitigate catastrophic forgetting is by regulating the storage and retrieval of memories in a sparsely-distributed, context-gated manner. As mentioned earlier, pyramidal neurons are not just simple thresholding units of synaptic inputs. They can also function as powerful, context-sensitive processors that allow distal contextual signals to gate feedforward information. Although the mechanisms underlying this process were not fully understood back then, Masse et al. (2018) [33] proposed the Context-dependent Gating (XdG) algorithm, which abstractly mimics dendritic gating in DNNs. XdG works by providing the network with task-specific, contextual signals that deactivate a percentage of neurons when trained on a task. This approach ensures that only sparse, mostly non-overlapping patterns of units are active for any given task, greatly reducing interference when trained on multiple tasks. More recently, Iyer et al. (2022) [34] achieved similar results in a bioinspired DNN, where each neuron comprised a variable number of context-receiving dendritic nodes. These were connected to a single input-receiving node, and a neuron could only be activated if it received both sensory information and contextual signals to at least one of its dendrites. This approach allowed the network to invoke minimally overlapping subnetworks for different tasks learned sequentially, whose activation is context-specific.

## Low-power neuromorphic computing

Neuromorphic computing, in general, aims to emulate the information processing and storage mechanisms of the brain. Compared to traditional von Neumann architectures, neuromorphic systems are significantly more efficient due to the close integration of computing and memory within each neuron [3,7]. This enables local processing and storage of information, reducing the need for frequent data transfers between different device components. In addition, biological neurons communicate sparsely with each other in an event-driven manner, further improving the efficiency of such systems. Since dendrites are indispensable for the above processes, dendritic mechanisms are expected to have merit in neuromorphic research. Below, we highlight studies providing tangible examples of how dendrites can help develop low-power, neuro-inspired hardware (**Figure 2b-c**).

Bhaduri et al (2018), provided a proof-of-concept that dendritic properties can be leveraged to perform data classification efficiently in hardware [35]. They developed a spiking neural classifier with non-linear dendrites that was trained using only binary synapses and a structural synaptic plasticity rule. This system achieved classification accuracy similar to conventional machine learning algorithms, requiring much fewer synaptic resources. Notably, since the hardware implementation of dendrites is way more compact than a neuron, their addition enhanced the computational power of the chip without requiring significant additional surface area. More recently, Gao et al. (2022) [36] developed a neuromorphic system based on Guerguiev et al.'s (2017) learning rule, mentioned earlier [17]. In contrast to backprop, which requires error gradients to be communicated across the entire network, the errors in this system are computed locally at the dendritic level. This reduces the system's complexity and enables data to be stored directly on the device, rather than off-chip memory. The reduced data flow between on-chip and off-chip memories was shown to greatly reduce the device's overall energy consumption.

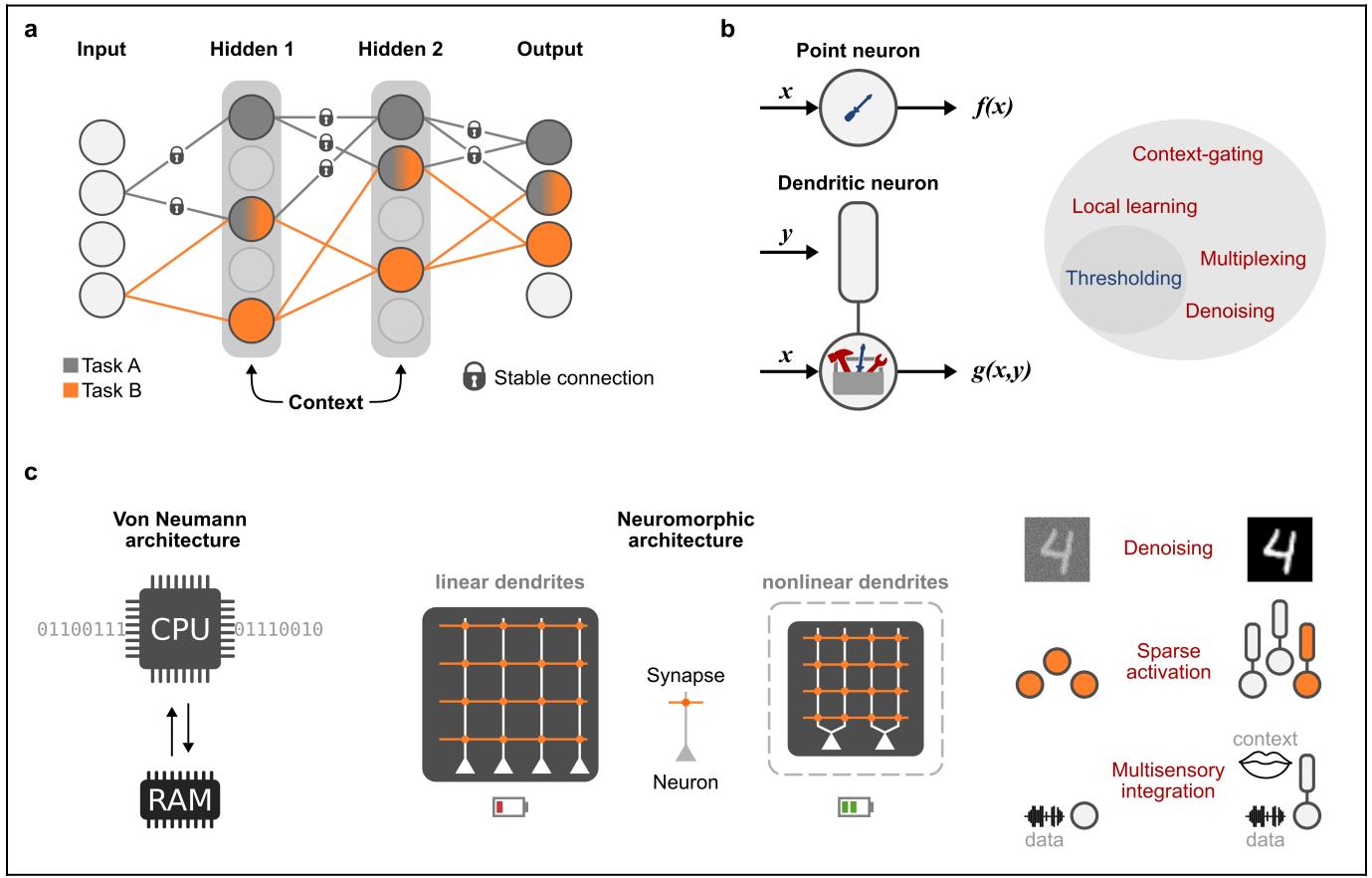

**Figure 2 | Dendritic mechanisms convey significant advantages DNNs and neuromorphic hardware**

**a)** Dendro-inspired algorithms can significantly increase the number of tasks a single network can learn, thus mitigating catastrophic forgetting. Synaptic weights that are most important for a learned task (gray) are made less plastic when a new task is learned, forcing the remaining synapses (orange) to be utilized. This allows for robust learning of each task, utilizing different subnetworks, thus greatly reducing memory interference.

**b)** At the conceptual level, a point neuron acts as a simple thresholding device. Adding active dendritic compartments can greatly expand the repertoire of computations that a single neuron can perform.

**c)** <u>Left</u>: Traditional Von Neumann architectures require frequent data transfers between a central processing unit (CPU) and a rapid access memory (RAM). <u>Middle</u>: Neuromorphic chips incorporating non-linear dendritic mechanisms can have more efficient surface area utilization and consume less power. Right: Example tasks performed by neuromorphic chips with non-linear dendrites demonstrating their computational advantages.

A fundamental limitation of point-neuron models is that they always process incoming information, regardless of whether it is essential or useful for a specific task. In DNNs this can cause a cascade of unnecessary neuronal activations that distract the network from a target goal, negatively impacting its performance and efficiency. To address this issue, Adeel et al. (2022) proposed a novel network architecture that utilizes two-point, context-sensitive neurons [37]. The innovation of this approach is that neurons integrate multimodal streams of information to infer the relevance of a given input (context). As a result, they are selectively activated only when the received information is relevant to the task at hand. To test this architecture, the authors employed a challenging audio-visual speech processing task that uses video information from lip movements to selectively amplify speech signals heard in noisy environments. Interestingly, the results showed similar performance compared to other state-of-the-art approaches but with substantially lower hardware power consumption. In this case, resource savings emanated from the ability of the network to extract relevant features at very early stages, leading to faster learning and sparser neuronal activation.

The latest advances in the field of memristors shed light on how dendritic mechanisms can be harnessed to create powerful and energy-efficient systems. Li and colleagues (2020 & 2022) demonstrated that incorporating dendritic nonlinearities into memristive devices can significantly enhance their performance and efficiency [38,39]. Adding active dendrites introduces an extra thresholding step in input processing that is optimal for denoising and maintaining sparse network activation. Notably, the proposed architecture achieved compelling performance when tested on challenging tasks, including noisy image classification and human motion recognition, at a fraction of the cost of modern CPUs or GPUs. A key advantage of memristive dendrites is that they enhance network performance by enabling richer neuronal dynamics. This approach is way more efficient regarding hardware cost and consumes less energy than simply scaling the network with additional neurons and synapses.

## Conclusion

The above studies emphasize the significant advantages of integrating dendritic properties into AI algorithms and neuromorphic hardware. These benefits are gaining widespread recognition, signaling an upcoming paradigm shift in machine learning and neuromorphic engineering, wherein dendrites will play a central role. More research in bio-inspired network architectures and learning rules is required to unleash the full potential of dendritic systems in solving intricate tasks, bringing artificial systems a step closer to the impressive capabilities of biological brains. The era of dendro-inspiration has begun, holding great promise for exciting and transformative advancements in the years to come.

## Conflict of interest

The authors declare no competing interests.

## Acknowledgments

We would like to thank Dr. Spyridon Chavlis and other members of the Poirazi lab for their valuable feedback on the manuscript. This work was supported by NIH (1R01MH124867-02), the European Commission (H2020-FETOPEN-2018-2019-2020-01, NEUREKA GA-863245, and H2020 MSCA ITN Project SmartNets GA-860949), and the Einstein Foundation Berlin, Germany (visiting fellowship to PP, EVF-2019-508).

## Declaration of Generative AI and AI-assisted technologies in the writing process

During the preparation of this work, the authors used ChatGPT (GPT-3.5) in order to improve the readability and language of the manuscript. After using this tool, the authors reviewed and edited the content as needed and take full responsibility for the content of the publication.

# References and recommended reading

Papers of particular interest published within the period of review have been highlighted as:
* of special interest
** of outstanding interest